\pdfoutput=1

\documentclass[11pt]{article}

\usepackage[]{ACL2023}

\usepackage{times}
\usepackage{latexsym}

\usepackage[T1]{fontenc}

\usepackage[utf8]{inputenc}

\usepackage{microtype}

\usepackage{inconsolata}
\usepackage{graphicx}
\usepackage{amsmath}

\DeclareMathOperator*{\argmin}{arg\,min}
\usepackage{amssymb}
\usepackage{booktabs}
\usepackage{algorithm}
\usepackage{algorithmic}
\usepackage{xcolor}
\usepackage{tabularx}
\usepackage{subcaption}
\usepackage{multirow}
\usepackage{pifont}
\usepackage{xspace}
\usepackage[normalem]{ulem} 
\usepackage{newunicodechar}
\newunicodechar{✓}{\ding{51}}
\newunicodechar{✗}{\ding{55}}

\newcommand{\Approach}[1]{DGSlow}

\title{White-Box Multi-Objective Adversarial Attack on Dialogue Generation}

\author{
Yufei Li,
\xspace    
Zexin Li,
\xspace   
Yingfan Gao,
\xspace 
Cong Liu
\\
University of California, Riverside\\ 
\texttt{\{yli927,zli536,ygao195,congl\}@ucr.edu}
}

\begin{document}

\maketitle

\begin{abstract}
Pre-trained transformers are popular in state-of-the-art dialogue generation~(DG) systems.
Such language models are, however, vulnerable to various adversarial samples as studied in traditional tasks such as text classification, which inspires our curiosity about their robustness in DG systems. 
One main challenge of attacking DG models is that perturbations on the current sentence can hardly degrade the response accuracy because the unchanged chat histories are also considered for decision-making.
Instead of merely pursuing pitfalls of performance metrics such as BLEU, ROUGE, we observe that crafting adversarial samples to force longer generation outputs benefits attack effectiveness---the generated responses are typically irrelevant, lengthy, and repetitive.
To this end, we propose a white-box multi-objective attack method called \textbf{\Approach{}}.
Specifically, \Approach{} balances two objectives---generation accuracy and length, via a gradient-based multi-objective optimizer and applies an adaptive searching mechanism to iteratively craft adversarial samples with only a few modifications. 
Comprehensive experiments\footnote{Our code is available at \url{https://github.com/yul091/DGSlow.git}} on four benchmark datasets demonstrate that \Approach{} could significantly degrade state-of-the-art DG models with a higher success rate than traditional accuracy-based methods. 
Besides, our crafted sentences also exhibit strong transferability in attacking other models.

\end{abstract}

\section{Introduction}

Pre-trained transformers have achieved remarkable success in dialogue generation (DG)~\cite{dialogpt,T5,roller-etal-2021-recipes}, e.g., the ubiquitous chat agents and voice-embedded chat-bots. 
However, such powerful models are fragile when encountering adversarial samples crafted by small and imperceptible perturbations~\cite{goodfellow2014explaining}.
Recent studies have revealed the vulnerability of deep learning in traditional tasks such as text classification~\cite{chen-etal-2021-multi,guo-etal-2021-gradient,zeng-etal-2021-openattack} and neural machine translation~\cite{zou-etal-2020-reinforced,zhang-etal-2021-crafting}. 
Nonetheless, investigating the robustness of DG systems has not received much attention.

Crafting DG adversarial samples is notably more challenging due to the conversational paradigm, where we can only modify the current utterance while the models make decisions also based on previous chat history~\cite{liu-etal-2020-impress}. 
This renders small perturbations even more negligible for degrading the output quality.
An intuitive adaptation of existing accuracy-based attacks, especially black-box methods~\cite{scpn,PWWS,zhang-etal-2021-crafting} that merely pursue pitfalls for performance metrics, cannot effectively tackle such issues.
Alternatively, we observed that adversarial perturbations forcing longer outputs are more effective against DG models, as longer generated responses are generally more semantic-irrelevant to the references.
Besides, such an objective is non-trivial because current large language models can handle and generate substantially long outputs.
This implies the two attacking objectives---generation accuracy and length, can somehow be correlated and jointly approximated.

To this end, we propose a novel attack method targeting the two objectives called \textbf{\Approach{}}, which produces semantic-preserving adversarial samples and achieves a higher attack success rate on DG models.
Specifically, we define two objective-oriented losses corresponding to the response accuracy and length.
Instead of integrating both objectives and applying human-based parameter tuning, which is inefficient and resource-consuming, we propose a gradient-based multi-objective optimizer to estimate an optimal Pareto-stationary solution~\cite{multi-objective-optimization}.
The derived gradients serve as indicators of the significance of each word in a DG instance.
Then we iteratively substitute those keywords using masked language modeling~(MLM)~\cite{devlin-etal-2019-bert} and validate the correctness of crafted samples.
The intuition is to maintain semantics and grammatical correctness with minimum word replacements~\cite{zou-etal-2020-reinforced,cheng-etal-2020-advaug}. 
Finally, we define a unique fitness function that considers both objectives for selecting promising crafted samples.
Unlike existing techniques that apply either greedy or random search, we design an adaptive search algorithm where the selection criteria are dynamically based on the current iteration and candidates' quality.
Our intuition is to avoid the search strapped in a local minimum and further improve efficiency.

We conduct comprehensive attacking experiments on three pre-trained transformers over four DG benchmark datasets to evaluate the effectiveness of our method.
Evaluation results demonstrate that \Approach{} overall outperforms all baseline methods in terms of higher attack success rate, better semantic preservance, and longer as well as more irrelevant generation outputs.
We further investigate the transferability of \Approach{} on different models to illustrate its practicality and usability in real-world applications.

Our main contributions are as follows:

\begin{itemize}
    \item To the best of our knowledge, we are the first to study the robustness of large language models in DG systems against adversarial attacks, and propose a potential way to solve such challenge by re-defining DG adversarial samples.
    \item Different from existing methods that only consider a single objective, e.g., generation accuracy, we propose multi-objective optimization and adaptive search to produce semantic-preserving adversarial samples that can produce both lengthy and irrelevant outputs.
    \item Extensive experiments demonstrate the superiority of \Approach{} to all baselines as well as the strong transferability of our crafted samples.  
\end{itemize}

\section{Dialogue Adversarial Generation}
\label{sec:dg settings}

Suppose a chat bot aims to model conversations between two persons.
We follow the settings \cite{liu-etal-2020-impress} where each person has a persona (e.g., $\boldsymbol{c}^{\mathcal{A}}$ for person $\mathcal{A}$), described with $L$ profile sentences $\left \{ c_1^{\mathcal{A}}, ..., c_L^{\mathcal{A}} \right \}$.
Person $\mathcal{A}$ chats with the other person $\mathcal{B}$ through a $N$-turn dialogue $(x_1^{\mathcal{A}}, x_1^{\mathcal{B}}, ..., x_N^{\mathcal{A}}, x_N^{\mathcal{B}})$, where $N$ is the number of total turns and $x_n^{\mathcal{A}}$ is the utterance that $\mathcal{A}$ says in $n$-th turn.
A DG model $f$ takes the persona $\boldsymbol{c}^{\mathcal{A}}$, the entire dialogue history until $n$-th turn $\boldsymbol{h}_n^\mathcal{A}=(x_1^{\mathcal{B}}, ..., x_{n-1}^{\mathcal{A}})$, and $\mathcal{B}$'s current utterance $x_n^{\mathcal{B}}$ as inputs, generates outputs $x_n^{\mathcal{A}}$ by maximizing the probability $p(x_n^{\mathcal{A}}|\boldsymbol{c}^{\mathcal{A}}, \boldsymbol{h}_n^\mathcal{A}, x_n^{\mathcal{B}})$. 
The same process applies for $\mathcal{B}$ to keep the conversation going.
In the following, we first define the optimization goal of DG adversarial samples and then introduce our multi-objective optimization followed by a search-based adversarial attack framework.

\subsection{Definition of DG Adversarial Samples}

In each dialogue turn $n$, we craft an utterance $x_n^{\mathcal{B}}$ that person $\mathcal{B}$ says to fool a bot targeting to mimic person $\mathcal{A}$. 
Note that we do not modify the chat history $\boldsymbol{h}_n^\mathcal{A}=(x_1^{\mathcal{B}}, ..., x_{n-1}^{\mathcal{A}})$, as it should remain unchanged in real-world scenarios.

Take person $\mathcal{B}$ as an example, an optimal DG adversarial sample in $n$-th turn is a utterance $x_n^{\mathcal{B}*}$:

\begin{equation}
\label{eq:define}
\begin{gathered}
x_n^{\mathcal{B}*} = \argmin_{\hat{x}_n^{\mathcal{B}}} M(x^{ref}_{n}, \hat{x}_n^{\mathcal{A}}) \\
s.t. \; \hat{x}^{\mathcal{A}}_n \equiv f(\boldsymbol{c}^{\mathcal{A}}, \boldsymbol{h}_n^\mathcal{A}, \hat{x}_n^{\mathcal{B}}) 
\wedge
\rho(x_n^{\mathcal{B}}, \hat{x}_n^{\mathcal{B}}) > \epsilon
\end{gathered}
\end{equation}

where $\rho(.)$ is a metric for measuring the semantic preservance, e.g., the cosine similarity between the original input sentence $x_n^{\mathcal{B}}$ and a crafted sentence $\hat{x}_n^{\mathcal{B}}$.
$\epsilon$ is the perturbation threshold.
$M(\cdot)$ is a metric for evaluating the quality of an output sentence $\hat{x}_n^{\mathcal{A}}$ according to a reference $x^{ref}_{n}$.
Existing work typically applies performance metrics in neural machine translation~(NMT), e.g., BLEU score~\cite{bleu}, ROUGE~\cite{rouge}, as a measurement of $M(\cdot)$. 
In this work, we argue the output length itself directly affects the DG performance, and generating longer output should be considered as another optimization objective.

Accordingly, we define \textit{Targeted Confidence} (TC) and \textit{Generation Length} (GL). 
TC is formulated as the cumulative probabilities regarding a reference $x_{n}^{ref}$ to present the accuracy objective, while GL is defined as the number of tokens in the generated output sentence regarding an input $\hat{x}_n^{\mathcal{B}}$ to reflect the length objective:

\begin{equation}
\label{eq:tc_gl}
  \left\{ \\
  \begin{array}{ll}
     \text{TC}(\hat{x}_n^{\mathcal{B}}) = \sum_{t}p_{\theta}(x_{n,t}^{ref}|\boldsymbol{c}^{\mathcal{A}}, \boldsymbol{h}_n^\mathcal{A}, \hat{x}_n^{\mathcal{B}}, x_{n,<t}^{ref})  &  \\
     \text{GL}(\hat{x}_n^{\mathcal{B}}) = |\hat{x}^{\mathcal{A}}_n| = |f(\boldsymbol{c}^{\mathcal{A}}, \boldsymbol{h}_n^\mathcal{A}, \hat{x}_n^{\mathcal{B}})|  & 
  \end{array}
  \right.
\end{equation}

\begin{figure}[]
    \centering
    \includegraphics[width=0.485\textwidth]{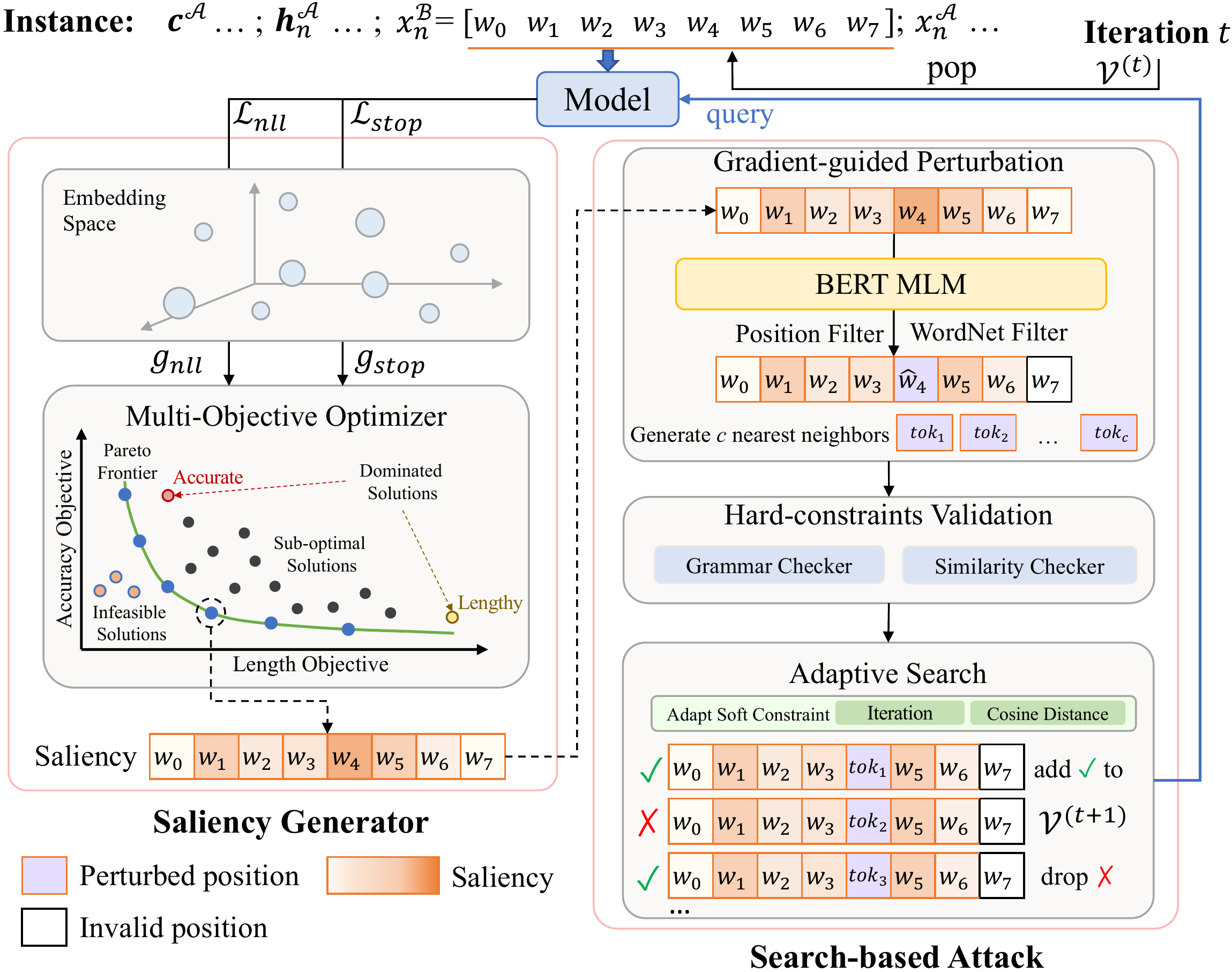}
    \caption{Illustration of our \Approach{} attack method. 
    In each iteration, the current adversarial utterance $\hat{x}_n^{\mathcal{B}}$, together with persona, chat history, and references, are fed into the model to obtain the word saliency via gradient descent.
    Then we mutate the positions with high word saliency and validate the correctness of the perturbed samples.
    The remaining samples query the model to calculate their fitness, and we select $k$ prominent candidates using adaptive search for the next iteration.
    }
    \label{fig:architecture}
\end{figure}

Based on our DG definition in Eq.~(\ref{eq:define}), we aim to craft adversarial samples that could produce small TC and large GL. 
To this end, we propose a white-box targeted DG adversarial attack that integrates multi-objective optimization and adaptive search to iteratively craft adversarial samples with word-level perturbations (see Figure~\ref{fig:architecture}).

\subsection{Multi-Objective Optimization}
\label{sec:MO}

Given a DG instance $(\boldsymbol{c}^{\mathcal{A}}, \boldsymbol{h}_n^\mathcal{A}, x_n^{\mathcal{B}}, x_n^{ref})$, an appropriate solution to produce lower TC is to minimize the log-likelihood~(LL) objective for decoding $x_n^{ref}$, i.e., the accumulated likelihood of next token $x_{n,t}^{ref}$ given previous tokens $x_{n,<t}^{ref}$:

\begin{equation}
    \mathcal{L}_{ll} = \sum_t \log p_{\theta}(x_{n,t}^{ref}|\boldsymbol{c}^{\mathcal{A}}, \boldsymbol{h}_n^\mathcal{A}, x_n^{\mathcal{B}}, x_{n,<t}^{ref})
\end{equation}

In another aspect, crafting adversarial samples with larger GL can be realized by minimizing the decoding probability of \textit{eos} token, which delays the end of decoding process to generate longer sequences. 
Intuitively, without considering the implicit Markov relationship in a DG model and simplifying the computational cost, we directly force an adversarial example to reduce the probability of predicting \textit{eos} token by applying the Binary Cross Entropy (BCE) loss:

\begin{equation}
\label{eq:eos_loss}
    \mathcal{L}_{eos} = \sum_{t} (l_t^{eos} - \mathbb{E}_{tok\sim p_t} l_t^{tok})
\end{equation}

where $l_t^{tok}$ is the logit at position $t$ regarding a predicted token $tok$, and $p_t$ is the decoding probability for the $t$-th token.
Furthermore, we penalize adversarial samples that deviate too much from the original sentence to preserve semantics:

\begin{equation}
\label{eq:per_loss}
    \mathcal{L}_{reg} = \max(0, \epsilon - \rho(x_n^{\mathcal{B}}, \hat{x}_n^{\mathcal{B}}))
\end{equation}

where $\rho$ and $\epsilon$ are semantic similarity and threshold as defined in Eq.~(\ref{eq:define}). 
We formulate the stop loss as a weighted sum of \textit{eos} loss and regularization penalty to represent the length objective:

\begin{equation}
\label{eq:stop_loss}
    \mathcal{L}_{stop} = \mathcal{L}_{eos} + \beta \mathcal{L}_{reg}
\end{equation}

where $\beta$ is a hyper-parameter that controls the penalty term's impact level. 
Considering that the log-likelihood loss $\mathcal{L}_{ll}$ and the stop loss $\mathcal{L}_{stop}$ may conflict to some extent as they target different objectives, we assign proper weights $\alpha_1$, $\alpha_2$ to each loss and optimize them based on the \textit{Multi-objective Optimization}~(MO) theorem~\cite{multi-objective-optimization}. 
Specifically, we aim to find a Pareto-stationary point by solving the Lagrange problem:

\begin{equation}
\begin{gathered}
\begin{pmatrix}
\hat{\alpha}_1^*\\ 
\hat{\alpha}_2^*\\ 
\lambda
\end{pmatrix}=(\mathcal{M}^{\top}\mathcal{M})^{-1}\mathcal{M}
\begin{bmatrix}
-\mathcal{G}\mathcal{G}^{\top}\boldsymbol{c}\\ 
1-\boldsymbol{e}^{\top}\boldsymbol{c}\\ 
\lambda
\end{bmatrix}
\\
s.t. \;
\mathcal{M}=
\begin{bmatrix}
\mathcal{G}\mathcal{G}^{\top} & \boldsymbol{e} \\
\boldsymbol{e}^{\top} & 0
\end{bmatrix}
\end{gathered}
\end{equation}

where $\mathcal{G}=[g_{ll}, g_{stop}]$, and $g_{ll}$, $g_{stop}$ are gradients derived from $\mathcal{L}_{ll}$, $\mathcal{L}_{stop}$ w.r.t. the embedding layer,
$\boldsymbol{e}=[1,1]$, $\boldsymbol{c}=[c_1,c_2]$ and $c_1$, $c_2$ are two boundary constraints $\alpha_1 \geq c_1, \alpha_2 \geq c_2$,
$\lambda$ is the Lagrange multiplier.
The final gradient is defined as the weighted sum of the two gradients $g=\hat{\alpha}_1^{*} \cdot g_{ll} + \hat{\alpha}_2^{*}\cdot  g_{stop}$.
Such gradients facilitate locating the significant words in a sentence for effective and efficient perturbations.

\subsection{Search-based Adversarial Attack}
\label{sec:BS}
We combine the multi-objective optimization with a search-based attack framework to iteratively generate adversarial samples against the DG model, as shown in the right part of Figure~\ref{fig:architecture}.
Specifically, our search-based attacking framework contains three parts---\textit{Gradient-guided Perturbation}~(GP) that substitutes words at significant positions, \textit{Hard-constraints Validation}~(HV) that filters out invalid adversarial candidates, and \textit{Adaptive Search}~(AS) that selects $k$ most prominent candidates based on different conditions for the next iteration.

\textbf{Gradient-guided Perturbation.} Let $x=[w_0,...,w_i, ..., w_n]$ be the original sentence where $i$ denotes the position of a word $w_i$ in the sentence.
During iteration $t$, for the current adversarial sentence $\hat{x}^{(t)}=[w_0^{(t)},...,w_i^{(t)}, ..., w_n^{(t)}]$, we first define \textit{Word Saliency} (WS)~\cite{li-etal-2016-visualizing} which is used to sort the positions whose corresponding word has not been perturbed.
The intuition is to skip the positions that may produce low attack effect so as to accelerate the search process.
In our DG scenario, WS refers to the significance of a word in an input sentence for generating irrelevant and lengthy output.
We quantified WS by average pooling the aforementioned gradient $g$ over the embedding dimension, and sort the positions according to an order of large-to-small scores.

For each position $i$, we define a candidate set $\mathbb{L}^{(t)}_{i}\in \mathbb{D}$ where $\mathbb{D}$ is a dictionary consisting of all words that express similar meanings to $w_i^{(t)}$, considering the sentence context.
In this work, we apply BERT masked language modeling~(MLM)~\cite{devlin-etal-2019-bert} to generate $c$ closest neighbors in the latent space.
The intuition is to generate adversarial samples that are more fluent compared to rule-based synonymous substitutions.
We further check those neighbors by querying the WordNet~\cite{miller1998wordnet} and filtering out antonyms of $w_i^{(t)}$  to build the candidate set.
Specifically, we first create a masked sentence $x_{m_i}^{(t)}=[w_0^{(t)},...,\text{[MASK]},...,w_n^{(t)}]$ by replacing $w_i^{(t)}$ with a [MASK] token.
Then, we craft adversarial sentences $\hat{x}_i^{(t+1)}$ by filling the $\text{[MASK]}$ token in $x_{m_i}^{(t)}$ with different candidate tokens $\hat{w}_i^{(t+1)}$.

\textbf{Hard-constraints Validation.}
The generated adversarial sentence $\hat{x}^{(t)}$ could be much different from the original $x$ after $t$ iterations.
To promise \textit{fluency}, we validate the number of grammatical errors in $\hat{x}^{(t)}$ using a Language Checker~\cite{myint_language_check}.
Besides, the adversarial candidates should also preserve enough semantic information of the original one. 
Accordingly, we encode $\hat{x}^{(t)}$ and $x$ using a universal sentence encoder~(USE)~\cite{use2018}, and calculate the cosine similarity between their sentence embeddings as their semantic similarity.
We record those generated adversarial candidates $\hat{x}^{(t)}$ whose 1) grammar errors are smaller than that of $x$ and 2) cosine similarities with $x$ are larger than a predefined threshold $\epsilon$, then put them into a set $\mathcal{V}^{(t)}$, which is initialized before the next iteration.

\textbf{Adaptive Search.}
For a DG instance $(\boldsymbol{c}^{\mathcal{A}}, \boldsymbol{h}_n^\mathcal{A}, \hat{x}_n^{\mathcal{B}}, x_n^{ref})$, we define a domain-specific \textit{fitness} function $\varphi$ which measures the preference for a specific adversarial $\hat{x}_n^{\mathcal{B}}$:

\begin{equation}
    \label{eq:fitness}
    \varphi(\hat{x}_n^{\mathcal{B}}) = \frac{|f(\boldsymbol{c}^{\mathcal{A}}, \boldsymbol{h}_n^\mathcal{A}, \hat{x}_n^{\mathcal{B}})|}{\sum_{t}p_{\theta}(x_{n,t}^{ref}|\boldsymbol{c}^{\mathcal{A}}, \boldsymbol{h}_n^\mathcal{A}, \hat{x}_n^{\mathcal{B}}, x_{n,<t}^{ref})}
\end{equation}

The fitness serves as a criteria for selecting $\hat{x}_n^{\mathcal{B}}$ that could produce larger GL and has lower TC with respect to the references $x_n^{ref}$, considering the persona $\boldsymbol{c}^{\mathcal{A}}$ and chat history $\boldsymbol{h}_n^{\mathcal{A}}$.

After each iteration, it is straightforward to select candidates using \textit{Random Search}~(RS) or \textit{Greedy Search}~(GS) based on candidates' fitness scores.
However, random search ignores the impact of an initial result on the final result, while greedy search neglects the situations where a local optimum is not the global optimum.
Instead, we design an adaptive search algorithm based on the iteration $t$ as well as the candidates' quality $q_t$.   
Specifically, $q_t$ is defined as the averaged cosine similarity between each valid candidate and the original input:

\begin{equation}
    q_t = \frac{\sum_{\hat{x}^{(t)}\in \mathcal{V}^{(t)}}cos(\hat{x}^{(t)}, x)}{|\mathcal{V}^{(t)}|}
\end{equation}

Larger $q_t$ means smaller perturbation effects. 
The search preference $\xi_t$ can be formulated as:

\begin{equation}
    \xi_t=\frac{(t-1)e^{q_t - 1}}{T-1}    
\end{equation}

where $T$ is the maximum iteration number.
Given $t=[1, ..., T]$ and $q_t\in [0,1]$, $\xi_t$ is also bounded in the range $[0,1]$.
We apply random search if $\xi_t$ is larger than a threshold $\delta$, and greedy search otherwise.
The intuition is to 1) find a prominent initial result using greedy search at the early stage (small $t$), and 2) avoid being strapped into a local minimum by gradually introducing randomness when there is no significant difference between the current adversarial candidates and the prototype (large $q_t$).
We select $k$ (beam size) prominent candidates in $\mathcal{V}^{(t)}$, where each selected sample serves as an initial adversarial sentence in the next iteration to start a new local search for more diverse candidates.
We keep track of the perturbed positions for each adversarial sample to avoid repetitive perturbations and further improve efficiency.

\begin{table*}
\centering
\resizebox{0.95\textwidth}{!}{
    \begin{tabular}{c|cccc|cccc|cccc}
    \hline
        \multirow{2}{*}{\textbf{Dataset}} & 
        \multicolumn{4}{c|}{\textbf{DialoGPT}} & 
        \multicolumn{4}{c|}{\textbf{BART}} 
        & \multicolumn{4}{c}{\textbf{T5}} \\
        & \textbf{GL} 
        & \textbf{BLEU}  
        & \textbf{ROU.} 
        & \textbf{MET.} 
        & \textbf{GL} 
        & \textbf{BLEU}  
        & \textbf{ROU.} 
        & \textbf{MET.} 
        & \textbf{GL} 
        & \textbf{BLEU}  
        & \textbf{ROU.} 
        & \textbf{MET.}
        \\
        \hline
        BST & 16.05 & 14.54 & 19.42 & 23.83 & 14.94 & 13.91 & 20.73 & 20.52 & 14.14 & 14.12 & 22.12 & 21.70 \\
        PC & 15.22 & 18.44 & 30.23 & 31.03 & 13.65 & 18.12 & 28.30 & 28.81 & 13.12 & 18.20 & 28.83 & 28.91 \\
        CV2 & 12.38 & 12.83 & 16.31 & 14.10 & 10.64 & 12.24 & 11.81 & 12.03 & 13.25 & 10.23 & 10.61 & 9.24 \\
        ED & 14.47 & 9.24 & 13.10 & 11.42 & 14.69 & 8.04 & 11.13 & 10.92 & 15.20 & 7.73 & 11.31 & 10.34 \\
        \hline
    \end{tabular}}
    \caption{Performance of three DG victim models in four benchmark datasets. 
    GL denotes the average generation output length. 
    ROU.(\%) and MET.(\%) are abbreviations for ROUGE-L and METEOR.}
    \label{tab: orig_res}
\end{table*}

\section{Experiments}

\subsection{Experimental Setup}

\begin{table}[]
    \centering
    \resizebox{0.35\textwidth}{!}{
    \begin{tabular}{lcc}
        \hline
        \textbf{Dataset} & \textbf{\#Dialogues} & \textbf{\#Utterances}  \\
        \hline
        BST & 4,819 & 27,018  \\
        PC & 17,878 & 62,442 \\
        CV2 & 3,495 & 22,397 \\
        ED & 36,660 & 76,673 \\
        \hline
    \end{tabular}}
    \caption{Statistics of the four DG datasets.
    }
    \label{tab:datasets}
    \vspace{-0.5cm}
\end{table}

\textbf{Datasets.} We evaluate our generated adversarial DG examples on four benchmark datasets, namely, Blended Skill Talk (BST)~\cite{bst}, \textsc{PersonaChat} (PC)~\cite{personachat},  
ConvAI2 (CV2)~\cite{convai2}, and Empathetic-Dialogues (ED)~\cite{rashkin-etal-2019-towards}.
For BST and PC, we use their annotated suggestions as the references $x_{n}^{ref}$ for evaluation.
For ConvAI2 and ED, we use the response $x_n^{\mathcal{A}}$ as the reference since no other references are provided.
Note that we ignore the persona during inference for ED, as it does not include personality information.
We preprocess all datasets following the DG settings (in Section~\ref{sec:dg settings}) where each dialogue contains $n$-turns of utterances. 
The statistics of their training sets are shown in Table~\ref{tab:datasets}.

\textbf{Victim Models.}
We aim to attack three pre-trained transformers, namely, DialoGPT~\cite{dialogpt}, 
BART~\cite{lewis-etal-2020-bart}, and T5~\cite{T5}.
DialoGPT is pre-trained for DG on Reddit dataset, based on autoregressive GPT-2 backbones~\cite{gpt2}.
The latter two are seq2seq Encoder-Decoders pre-trained on open-domain datasets. 
Specifically, we use the HuggingFace pre-trained models---\textit{dialogpt-small}, \textit{bart-base}, and \textit{t5-small}.
The detailed information of each model can be found in Appendix~\ref{sec:additional_results}.
We use Byte-level BPE tokenization~\cite{gpt2} pre-trained on open-domain datasets, as implemented in HuggingFace tokenizers.
To meet the DG requirements, we also define two additional special tokens, namely, [PS] and [SEP].
[PS] is added before each persona to let the model be aware of the personality of each person.
[SEP] is added between each utterance within a dialogue so that the model can learn the structural information within the chat history.

\textbf{Metrics.}
We evaluate attack methods considering 1) the generation accuracy of adversarial samples 2) the generation length~(GL) of adversarial samples, and 3) the attack success rate~(ASR).
Specifically, the generation accuracy of adversarial samples are measured by performance metrics such as BLEU~\cite{bleu}, ROUGE-L~\cite{rouge, li-etal-2022-share} and METEOR~\cite{meteor} which reflect the correspondence between a DG output and references.    
We define ASR as:

\begin{equation}
\begin{gathered}
    \text{ASR} =
    \frac{\sum_i^N\textbf{1}[cos(x, \hat{x}) > \epsilon 
    \wedge
    E(y, \hat{y}) > \tau]}{N}
    \\
    s.t. \; E(y, \hat{y})=M(y, y_{ref})-M(\hat{y}, y_{ref})
\end{gathered}
\end{equation}

where $cos(.)$ denotes the cosine similarity between embeddings of original input $x$ and crafted input $\hat{x}$.
$M(\cdot, \cdot)$ is the average score of the three accuracy metrics.
An attack is successful if the adversarial input can induce a more irrelevant ($>\tau$) output and it preserves enough semantics ($>\epsilon$) of the original input.
Details of the performance of victim models are listed in Table~\ref{tab: orig_res}.

\textbf{Baselines.}
We compare against 5 recent white-box attacks and adapt their attacking strategy to our DG scenario, including four accuracy-based attacks:
1)~\textbf{FD}~\cite{MSH16-FD} conducts a standard gradient-based word substitution for each word in the input sentence,
2)~\textbf{HotFlip}~\cite{ebrahimi-etal-2018-hotflip} proposes adversarial attacks based on both word and character-level substitution using embedding gradients, 
3)~\textbf{TextBugger}~\cite{LiJDLW19-textbugger} proposes a greedy-based word substitution and character manipulation strategy to conduct the white-box adversarial attack against DG model, 
4)~\textbf{UAT}~\cite{wallace-etal-2019-universal} proposes word or character manipulation based on gradients. Specifically, its implementation relies on prompt insertion, which is different from most other approaches.
And one length-based attack \textbf{NMTSloth}~\cite{chen2022nmtsloth}, which is a length-based attack aiming to generate adversarial samples to make the NMT system generate longer outputs. It's a strong baseline that generates sub-optimal length-based adversarial samples even under several constraints.

For all baselines, we adapt their methodologies to DG scenarios, where the input for computing loss contains both the current utterance, and other parts of a DG instance including chat history, persona or additional contexts. 
Specifically, we use TC as the optimization objective (i.e., $\mathcal{L}_{ll}$) for all the baselines except NMTSloth which is a seq2seq attack method, and apply gradient descent to search for either word or character substitutions.

\textbf{Hyper-parameters.} 
For our DG adversarial attack, the perturbation threshold $\epsilon$ are performance threshold $\tau$ are set to 0.7 and 0 for defining a valid adversarial example. 
For multi-objective optimization, the regularization weight $\beta$ is set to 1 and the two boundaries $c_1$ and $c_2$ are set to 0 for non-negative constraints.
We use the Hugging face pre-trained \textit{bert-large-cased} model for MLM and set the number of candidates $c$ as 50 for mutation.
For adaptive search, we set the preference threshold $\delta$ as 0.5 and beam size $k$ as 2.
Our maximum number of iterations is set to 5, meaning that our modification is no more than 5 words for each sentence.
Besides, we also restrict the maximum query number to 2,000 for all attack methods.
For each dataset, we randomly select 100 dialogue conversations (each conversation contains 5$\sim$8 turns) for testing the attacking effectiveness.

\subsection{Overall Effectiveness}

\begin{table*}[htp]
\centering
\resizebox{0.99\textwidth}{!}{
    \begin{tabular}{c|c|ccccc|ccccc|ccccc}
    \hline
        \multirow{2}{*}{\textbf{Dataset}} & 
        \multirow{2}{*}{\textbf{Method}} &
        \multicolumn{5}{c|}{\textbf{DialoGPT}} &
        \multicolumn{5}{c|}{\textbf{BART}} &
        \multicolumn{5}{c}{\textbf{T5}} \\
        & & 
        \textbf{GL} & \textbf{BLEU} & 
        \textbf{ROU.} & 
        \textbf{ASR} &
        \textbf{Cos.} &
        \textbf{GL} & \textbf{BLEU}  & 
        \textbf{ROU.} & 
        \textbf{ASR} &
        \textbf{Cos.} &
        \textbf{GL} & \textbf{BLEU}  & 
        \textbf{ROU.} & 
        \textbf{ASR} &
        \textbf{Cos.} \\
        \hline
        \multirow{6}{*}{BST} & FD & 16.70 & 13.74 & 18.31 & 39.29 & 0.79 & 16.60 & 12.74 & 18.62 & 25.14 & 0.88 & 14.74 & 13.30 & 21.42 & 17.14 & 0.90 \\
        & HotFlip & 16.13 & 14.12 & 19.24 & 30.36 & 0.81 & 16.86 & 12.82 & 18.70 & 22.86 & 0.89 & 14.90 & 13.01 & 20.74 & 19.43 & 0.90 \\
        & TextBugger & 15.36 & 14.44 & 19.94 & 37.50 & 0.86 & 17.01 & 12.50 & 18.82 & 28.57 & 0.88 & 14.79 & 13.61 & 20.73 & 18.86 & 0.91 \\
        & UAT & 16.39 & 14.49 & 19.06 & 35.71 & \textbf{0.90} & 19.13 & 11.37 & 19.06 & 29.14 & \textbf{0.92} & 16.03 & 13.41 & 21.42 & 27.43 & 0.92 \\
        & NMTSloth & 22.23 & 13.20 & 18.65 & 55.36 & 0.78 & \textbf{23.74} & 9.60 & 17.91 & 42.45 & 0.84 & 27.31 & 9.49 & 18.37 & 48.57 & 0.85 \\
        & \Approach{} & \textbf{25.54} & \textbf{9.14} & \textbf{17.03} & \textbf{71.43} & \textbf{0.90} & 23.50 & \textbf{8.39} & \textbf{16.37} & \textbf{48.00} & \textbf{0.92} & \textbf{28.69} & \textbf{9.11} & \textbf{15.82} & \textbf{57.14} & \textbf{0.93} \\
        \hline
        \multirow{6}{*}{PC} & FD & 17.27 & 17.13 & 30.22 & 36.67 & 0.79 & 17.20 & 15.71 & 26.90 & 46.55 & 0.79 & 14.54 & 16.34 & 27.69 & 33.62 & 0.82 \\
        & HotFlip & 17.22 & 17.74 & 28.81 & 56.67 & 0.79 & 17.51 & 15.01 & 26.53 & 57.76 & 0.77 & 15.97 & 15.31 & 27.20 & 43.10 & 0.81 \\
        & TextBugger & 17.93 & 17.42 & 30.51 & 41.67 & 0.84 & 18.08 & 14.32 & 26.91 & 57.76 & 0.80 & 14.73 & 15.81 & 27.60 & 43.10 & 0.86 \\
        & UAT & 11.35 & 17.54 & 30.52 & 53.33 & \textbf{0.87} & 17.91 & 14.83 & 25.84 & 61.21 & \textbf{0.89} & 15.62 & 16.24 & 28.27 & 36.21 & 0.81 \\
        & NMTSloth & 22.01 & 16.39 & 28.79 & 66.67 & 0.73 & 29.09 & \textbf{8.96} & 21.49 & 95.69 & 0.58 & 30.37 & 8.87 & 16.66 & 87.93 & 0.65 \\
        & \Approach{} & \textbf{25.72} & \textbf{15.68} & \textbf{27.77} & \textbf{70.00} & 0.86 & \textbf{31.94} & 9.32 & \textbf{20.50} & \textbf{96.55} & \textbf{0.89} & \textbf{32.17} & \textbf{8.86} & \textbf{15.38} & \textbf{90.33} & \textbf{0.86} \\
        \hline
        \multirow{6}{*}{CV2} & FD & 15.74 & 12.54 & 14.33 & 38.10 & 0.78 & 12.30 & 10.81 & 10.52 & 20.13 & 0.88 & 13.97 & 9.91 & 10.62 & 16.78 & \textbf{0.90} \\
        & HotFlip & 16.38 & 13.33 & 15.21 & 33.33 & \textbf{0.81} & 13.46 & 10.50 & 10.41 & 32.89 & 0.86 & 14.03 & 9.63 & 10.12 & 26.17 & 0.86 \\
        & TextBugger & 12.93 & 12.83 & 14.71 & 40.48 & 0.80 & 12.70 & 10.82 & 10.12 & 34.90 & 0.87 & 15.00 & 9.62 & 10.11 & 27.52 & 0.87 \\
        & UAT & 14.36 & 12.94 & 15.79 & 42.86 & 0.80 & 13.50 & 10.61 & 10.23 & 33.56 & \textbf{0.88} & 15.17 & 9.21 & 10.11 & 30.20 & 0.85 \\
        & NMTSloth & 20.79 & 12.34 & 15.49 & 61.90 & 0.74 & 23.01 & 7.91 & 9.11 & 52.35 & 0.73 & 21.27 & 8.79 & 9.58 & 51.68 & 0.72 \\
        & \Approach{} & \textbf{28.54} & \textbf{11.70} & \textbf{13.71} & \textbf{64.29} & \textbf{0.81} & \textbf{23.84} & \textbf{6.51} & \textbf{8.34} & \textbf{56.61} & 0.87 & \textbf{22.32} & \textbf{7.74} & \textbf{8.43} & \textbf{53.02} & 0.88 \\
        \hline
        \multirow{6}{*}{ED} & FD & 15.00 & 9.03 & 12.62 & 41.82 & 0.75 & 19.66 & 6.54 & 10.44 & 44.26 & 0.76 & 16.66 & 7.41 & 11.30 & 32.79 & 0.79 \\
        & HotFlip & 17.69 & 8.71 & 12.92 & 40.74 & 0.78 & 21.38 & 6.71 & 10.74 & 67.21 & 0.70 & 17.30 & 7.03 & 10.81 & 37.70 & 0.80 \\
        & TextBugger & 14.66 & 9.01 & 12.73 & 40.00 & 0.89 & 22.26 & 6.03 & 8.82 & 70.49 & 0.78 & 17.11 & 7.12 & 10.23 & 47.54 & 0.81 \\
        & UAT & 15.33 & \textbf{8.64} & 13.03 & 52.73 & 0.87 & 20.72 & 6.41 & 11.12 & 50.82 & \textbf{0.82} & 17.30 & 7.24 & 10.43 & 42.62 & 0.89 \\
        & NMTSloth & 23.76 & 8.98 & 13.83 & 65.45 & 0.87 & 29.98 & 4.51 & 9.32 & 86.89 & 0.78 & 35.90 & 4.49 & 7.98 & 90.16 & 0.80 \\
        & \Approach{} & \textbf{24.72} & 8.93 & \textbf{12.12} & \textbf{69.81} & \textbf{0.90} & \textbf{34.28} & \textbf{4.22} & \textbf{8.11} & \textbf{98.36} & \textbf{0.82} & \textbf{38.82} & \textbf{4.02} & \textbf{6.10} & \textbf{94.16} & \textbf{0.92} \\
        \hline
    \end{tabular}}
    \caption{Evaluation of attack methods on three victim models in four DG benchmark datasets. 
    GL denotes the average generation output length. Cos. denotes the cosine similarity between original and adversarial sentences. ROU. (\%) denotes ROUGE-L. \textbf{Bold} numbers mean the best metric values over the six methods.}
    \label{tab: attack_res}
\end{table*}

Table~\ref{tab: attack_res} shows the GL, two accuracy metrics (METEOR results are in Appendix~\ref{sec:additional_results}), ASR and cosine results of all attack methods.
We observe that NMTSloth and our \Approach{} can produce much longer outputs than the other four baselines.
Accordingly, their attacking effectiveness regarding the output accuracy, i.e., BLEU and ROUGE-L, and ASR scores are much better than the four accuracy-based methods, proving the correctness of our assumption that adversarial samples forcing longer outputs also induce worse generation accuracy.
Though NMTSloth can also generate lengthy outputs as \Approach{} does, our method still achieves better ASR, accuracy scores and cosine similarity, demonstrating that our multi-objective optimization further benefits both objectives.
Moreover, our method can promise semantic-preserving perturbations while largely degrading the model performance, e.g., the cosine similarity of \Approach{} is at the top-level with baselines such as UAT and TextBugger.
This further proves our gradient-based word saliency together with the adaptive search can efficiently locate significant positions and  realize maximum attacking effect with only a few modifications. 

\begin{figure}[]
    \centering    \includegraphics[width=0.43\textwidth]{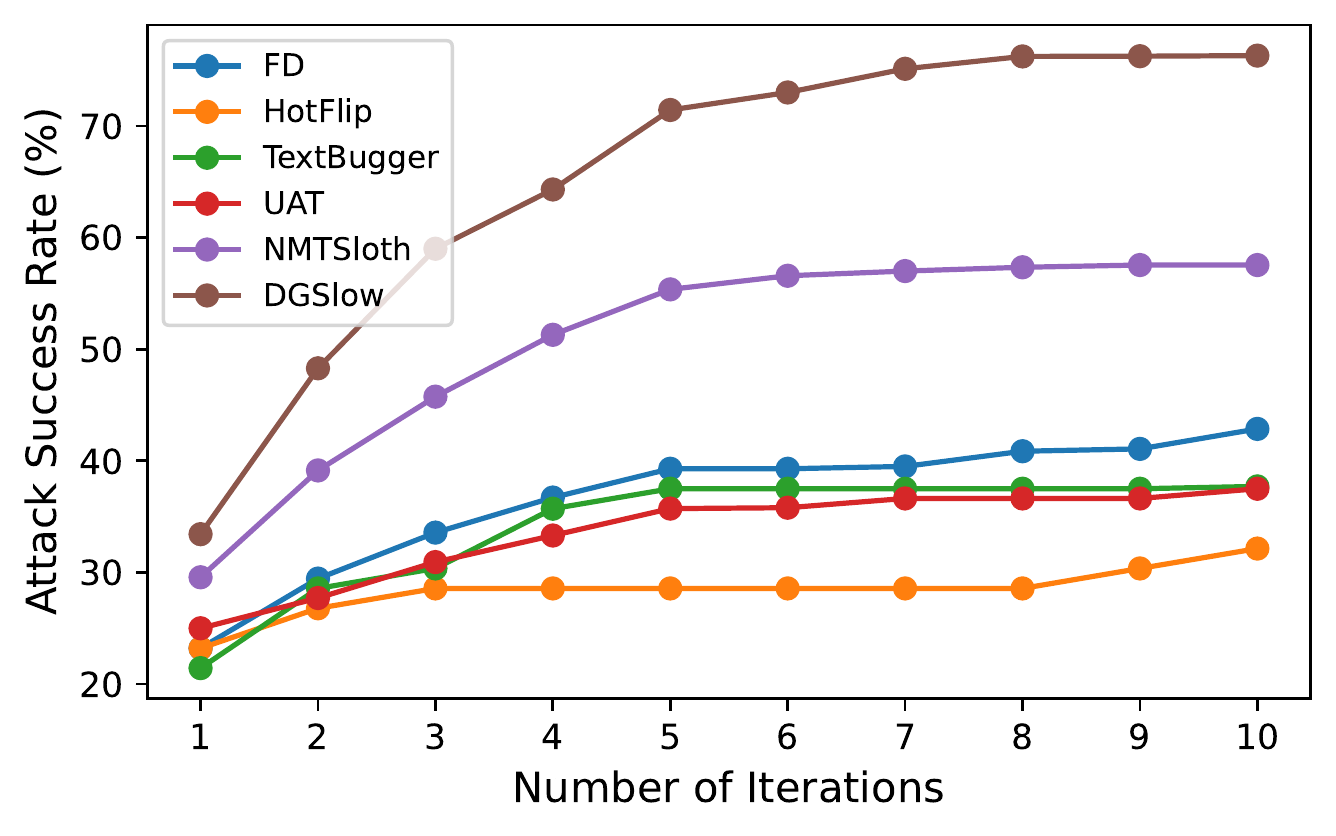}
    \caption{
    ASR vs. number of iterations in BST when attacking DialoGPT. 
    \Approach{} significantly outperforms all baselines.
    }
    \label{fig:asr_dialogpt}
    \vspace{-0.5cm}
\end{figure}

\begin{table}[]
    \centering
    \resizebox{0.43\textwidth}{!}{
    \begin{tabular}{c|ccccc}
    \hline
        \multirow{2}{*}{\textbf{Metric}} & \multicolumn{5}{c}{\textbf{Beam Size} $\boldsymbol{k}$}  \\
        \cline{2-6}
         & 1 & 2 & 3 & 4 & 5 \\
         \hline
         GL & 15.93 & 17.94 & 18.91 & 18.81 & 19.15 \\
         ASR & 46.98 & 47.99 & 48.32 & 48.65 & 49.32 \\
         BLEU & 13.06 & 12.93 & 11.27 & 10.90 & 9.03 \\
         \hline
    \end{tabular}}
    \caption{GL, ASR and BLEU vs. Beam size. In general, \Approach{} can produce adversarial samples that induce longer and more irrelevant outputs as the selected number of candidates after each iteration increases.}
    \label{tab:beam size}
\end{table}

\begin{table}[]
    \centering
    \resizebox{0.47\textwidth}{!}{\begin{tabular}{c|cccccc}
    \hline
        \textbf{Method} 
        & \textbf{MO} 
        & \textbf{CF} 
        & \textbf{BST} 
        & \textbf{PC} 
        & \textbf{CV2}
        & \textbf{ED}
        \\
        \hline
        RS & ✗ & ✗ & 30.29 & 61.21 & 30.87 & 52.46 \\
        GS & ✗ & ✗ & 46.29 & 85.69 & 48.99 & 86.89 \\
        \Approach{}$_1$ & ✗ & ✗ & 46.33 & 88.34 & 50.68 & 89.51 \\
        \Approach{}$_2$ & ✗ & ✓ & \textbf{48.33} & 90.16 & 49.65 & 90.25 \\
        \Approach{}$_3$ & ✓ & ✗ & 46.29 & 92.24 & 52.39 & 92.38 \\
        \Approach{} & ✓ & ✓ & 48.00 & \textbf{96.55} & \textbf{56.61} & \textbf{98.36} \\
        \hline
    \end{tabular}}
    \caption{Ablation study for ASR (\%) on BART with controllable components. 
    RS denotes random search. 
    GS denotes greedy search. 
    MO denotes multi-objective optimization. CF denotes combined fitness function.
    }
    \label{tab:ablation_study}
\end{table}

\begin{table}[]
    \centering
    \resizebox{0.48\textwidth}{!}{
    \begin{tabular}{cc|ccccc}
    \hline
        \textbf{Transfer} & \textbf{Victim} & \textbf{GL} & \textbf{BLEU} & \textbf{ROU.} &\textbf{MET.} & \textbf{ASR} \\
        \hline
        \multirow{2}{*}{DialoGPT} & BART & 20.35 & 8.53 & 10.79 & 8.68 & 55.81 \\
        & T5 & 19.02 & 9.18 & 10.91 & 8.66 & 47.50  \\
        \hline
        \multirow{2}{*}{BART} & DialoGPT & 25.73 & 7.84 & 10.67 & 10.90 & 67.27 \\
        & T5 & 24.71 & 7.91 & 10.03 & 10.92 & 63.93 \\
        \hline
        \multirow{2}{*}{T5} & DialoGPT & 23.89 & 7.70 & 11.28 & 10.33 & 47.27 \\
        & BART & 24.20 & 7.72 & 11.22 & 10.31 & 52.46 \\
        \hline
    \end{tabular}}
    \caption{Transfer attack results of adversarial samples in ED. 
    Victim denotes the model attacked by \Approach{} to generate adversarial samples. Transfer denotes the model that is tested by those crafted samples.
    }
    \label{tab:local_transfer}
\end{table}

\begin{table*}[]
    \centering
    \resizebox{0.97\textwidth}{!}{%
    
    \begin{tabular}{ll}
    \toprule
    \multicolumn{2}{l}{\textbf{Persona} $\boldsymbol{c}^{\mathcal{A}}$: I talked a lot in IRC.} \\
    \multicolumn{2}{l}{\textbf{Chat history} $\boldsymbol{h}$: } \\
    \multicolumn{2}{l}{[PERSON $\mathcal{B}$] You seem to know a lot about it. I chose the topic because I don't know anything about it.} \\
    \multicolumn{2}{l}{[PERSON $\mathcal{A}$] Yeah it's the chat process that works on a client/server model. It's a network chat. Do you want to know more?} \\
    
    \midrule
    \multicolumn{2}{l}{[$x_2^{\mathcal{B}}\rightarrow \hat{x}_2^{\mathcal{B}}$] Not really. Let's \textcolor{red}{\sout{talk}} \textcolor{blue}{\textit{think}} about food. What do you like to eat? I \textcolor{red}{\sout{love}} \textcolor{blue}{\textit{like}} fish.} \\
    \multicolumn{2}{l}{[$x_2^{\mathcal{A}}$] I love fish too! What is your favorite kind? I like pasta, steak, fish tacos etc.} \\
    \multicolumn{2}{l}{\begin{tabular}[c]{@{}l@{}}[$\hat{x}_2^{\mathcal{A}}$] I like to eat fish too. What is your favorite kind? I like pasta, filipino, steak, etc. I talk a lot on IRC and it is fun to learn\\about it with some other guys .\end{tabular}} \\

    \midrule
    \multicolumn{2}{l}{[$x_3^{\mathcal{B}}\rightarrow \hat{x}_3^{\mathcal{B}}$] I \textcolor{red}{\sout{eat}} \textcolor{blue}{\textit{take}} pretty much only fish. My parents do too, and they're both over 6 feet. Probably \textcolor{red}{\sout{cause of}} \textcolor{blue}{\textit{due to}} the fish.} \\
    \multicolumn{2}{l}{[$x_3^{\mathcal{A}}$] LOL, they're both over 6 feet! I can't imagine being that tall.} \\
    \multicolumn{2}{l}{\begin{tabular}[c]{@{}l@{}}[$\hat{x}_3^{\mathcal{A}}$] LOL.  Do you have a lot of fish, too?  My parents are over meaning feet. LOL. I don't know what they do due to the fish\\LOL. Do you guys like to talk a lot on IRC? \end{tabular}} \\

    \midrule
    \multicolumn{2}{l}{[$x_4^{\mathcal{B}}\rightarrow \hat{x}_4^{\mathcal{B}}$] I love salmon. \textcolor{red}{\sout{Sear}} \textcolor{blue}{\textit{Cook}} it with \textcolor{red}{\sout{some}} \textcolor{blue}{\textit{little}} rosemary, lots of butter, and some lemon.} \\
    \multicolumn{2}{l}{[$x_4^{\mathcal{A}}$] That's cool. I'm not sure what to eat, I'm not a big fish fan.} \\
    \multicolumn{2}{l}{\begin{tabular}[c]{@{}l@{}}[$\hat{x}_4^{\mathcal{A}}$] That sounds wonderful - what do you like for side dishes?  I eat lots of veggies', like asparagus fried with olive oil.\end{tabular}} \\

    \bottomrule
    \end{tabular}}
    \caption{\Approach{} crafts input sentences that cause DialoGPT to generate lengthy, irrelevant outputs. 
    \textcolor{blue}{\textit{Italics}} and \textcolor{red}{\sout{strike through}} denote added and removed tokens, respectively.}
    \label{tab:dialogue_adversary}
\end{table*}

\textbf{Attack Efficiency.} 
Figure~\ref{fig:asr_dialogpt} shows all attack methods' ASR in BST when attacking DialoGPT under the restriction of maximum iteration numbers.
Reminder results for the other two models can be found in Appendix~\ref{sec:additional_results}.
We observe that our attack significantly outperforms all accuracy-based baseline methods under the same-level of modifications, demonstrating the efficiency of length-based approach.
Furthermore, \Approach{} can achieve better ASR than NMTSloth, proving the practicality of our multi-objective optimization and adaptive search in real-world DG situations.

\textbf{Beam Size.}
We further evaluate the impact of the remaining number of prominent candidates $k$ (after each iteration) on the attack effectiveness, as shown in Table~\ref{tab:beam size}.
We observe that larger $k$ leads to overall longer GL, larger ASR and smaller BLEU, showing that as more diverse candidates are considered in the search space, \Approach{} is benefited by the adaptive search for finding better local optima.

\subsection{Ablation Study}

We exhibit the ablation study of our proposed \Approach{} algorithm in Table~\ref{tab:ablation_study}.
Specifically, if MO is not included, we only use gradient $g_{stop}$ derived from $\mathcal{L}_{stop}$ for searching candidates.
If CF is not included, we use $\varphi '(\hat{x}_n^{\mathcal{B}}) = \text{GL}(\hat{x}_n^{\mathcal{B}})$ as the fitness function, meaning we only select candidates that generate the longest output but ignore the quality measurement.
We observe that: 
1)~Greedily selecting candidates with highest fitness is more effective than random guess, e.g., the ASR of GS are much higher than those of RS; 
2)~Our adaptive search, i.e., \Approach{}$_1$, makes better choices when selecting candidates compared to RS and GS;
3)~Modifying the fitness function by considering both TC and GL, i.e., \Approach{}$_2$, can slightly improve overall ASR over \Approach{}$_1$;
4)~Only using multi-objective optimization, i.e., \Approach{}$_3$, can produce better attack results compared to only modifying the fitness.

\subsection{Transferability}

We evaluate the transferability of adversarial samples generated by our method on each model in ED with the other two as the victim models.
From Table~\ref{tab:local_transfer}, we observe that our \Approach{} can craft adversarial samples with decent transferability, e.g., the ASR are generally above 50\% , and the corresponding accuracy scores, e.g., BLEU, all decrease compared to those produced by original samples.
We believe it is because \Approach{} perturbs the sentence based on both accuracy and output length objectives, ensuring adversarial samples to capture more common vulnerabilities of different victim models than single objective based methods.

\subsection{Case Study}

We visualize three adversarial samples generated by \Approach{}, in Table~\ref{tab:dialogue_adversary}, which can effectively attack the DialoGPT model.
It shows that by replacing only several tokens with substitutions presenting similar meanings and part-of-speech tags, our method can induce the model to generate much longer, more irrelevant sequences $\hat{x}_n^{\mathcal{A}}$ compared to the original ones $x_n^{\mathcal{A}}$.
Such limited perturbations also promise the readability and semantic preservance of our crafted adversarial samples.

\section{Related Work}

\subsection{Adversarial Attack}

Various existing adversarial techniques raise great attention to model robustness in deep learning community~\cite{MSH16-FD,ebrahimi-etal-2018-hotflip,LiJDLW19-textbugger,wallace-etal-2019-universal,chen2022nmtsloth,ren-etal-2019-generating,zhang-etal-2021-crafting,li-etal-2020-bert-attack,li2023sibling}. 
Earlier text adversarial attacks explore character-based perturbations as they ignore out-of-vocabulary as well as grammar constraints, and are straightforward to achieve adversarial goals~\cite{DBLP:conf/iclr/BelinkovB18,ebrahimi-etal-2018-adversarial}. 
More recently, few attacks works focus on character-level~\cite{le-etal-2022-perturbations} since it's hard to generate non-grammatical-error adversarial samples without human study. 
Conversely, sentence-level attacks best promise grammatical correctness~\cite{chen-etal-2021-multi,scpn} but yield a lower attacking success rate due to change in semantics. 
Currently, it is more common to apply word-level adversarial attacks based on word substitutions, additions, and deletions~\cite{ren-etal-2019-generating,zou-etal-2020-reinforced,zhang-etal-2021-crafting,wallace-etal-2020-imitation,chen-etal-2021-multi}. 
Such strategy can better trade off semantics, grammatical correctness, and attack success rate.

Besides, a few researches focus on crafting attacks targeted to 
seq2seq tasks.
For example, NMTSloth~\cite{chen2022nmtsloth} targets to forcing longer translation outputs of an NMT system, while Seq2sick~\cite{Seq2Sick} and \cite{michel-etal-2019-evaluation} aim to degrade generation confidence of a seq2seq model.
Unlike previous works that only consider single optimization goal, we propose a new multi-objective word-level adversarial attack against DG systems which are challenging for existing methods.
We leverage the conversational characteristics of DG and redefine the attacking objectives to craft adversarial samples that can produce lengthy and irrelevant outputs.  

\subsection{Dialogue Generation}

Dialogue generation is a task to understand natural language inputs and produce human-level outputs, e.g., back and forth dialogue with a conversation agent like a chat bot with humans.
Some common benchmarks for this task include \textsc{PersonaChat}~\cite{personachat}, \textsc{FusedChat}~\cite{fusedchat}, Blended Skill Talk~\cite{bst}, ConvAI2~\cite{convai2}, Empathetic Dialogues~\cite{empathetic_dialogue}.
A general DG instance contains at least the chat history until the current turn, which is taken by a chat bot in structure manners to generate responses.
Recent DG chat bots are based on pre-trained transformers, including GPT-based language models such as DialoGPT~\cite{dialogpt}, PersonaGPT~\cite{personagpt}, and seq2seq models such as BlenderBot~\cite{roller-etal-2021-recipes}, T5~\cite{T5}, BART~\cite{lewis-etal-2020-bart}.
These large models can mimic human-like responses and even incorporate personalities into the generations if the user profile (persona) or some other contexts are provided.

\section{Conclusions}

In this paper, we propose \Approach{}---a white-box multi-objective adversarial attack that can effectively degrade the performance of DG models.
Specifically, \Approach{} targets to craft adversarial samples that can induce long and irrelevant outputs.
To fulfill the two objectives, it first defines two objective-oriented losses and applies a gradient-based multi-objective optimizer to locate key words for higher attack success rate. 
Then, \Approach{} perturbs words with semantic-preserving substitutions and selects promising candidates to iteratively approximate an optima solution.
Experimental results show that \Approach{} achieves state-of-the-art results regarding the attack success rate, the quality of adversarial samples, and the DG performance degradation.
We also show that adversarial samples generated by \Approach{} on a model can effectively attack other models, proving the practicability of our attack in real-world scenarios.

\section*{Limitations}

\noindent \textbf{Mutation.} 
We propose a simple but effective gradient-based mutation strategy. 
More complex mutation methods can be integrated into our framework to further improve attacking effectiveness. 

\noindent \textbf{Black-box Attack.} \Approach{} is based on a white-box setting to craft samples with fewer query times, but it can be easily adapted to black-box scenarios by using a non-gradient search algorithm, e.g., define word saliency based on our fitness function and do greedy substitutions.

\noindent \textbf{Adversarial Defense. } We do not consider defense methods in this work. 
Some defense methods, e.g.,  adversarial training and input denoising, may be able to defend our proposed \Approach{}.
Note that our goal is to pose potential threats by adversarial attacks and reveal the vulnerability of DG models, thus motivating the research of model robustness.

\section*{Ethics Statement}

In this paper, we design a multi-objective white-box attack against DG models on four benchmark datasets.
We aim to study the robustness of state-of-the-art transformers in DG systems from substantial experimental results and gain some insights about explainable AI. 
Moreover, we explore the potential risk of deploying deep learning techniques in real-world DG scenarios, facilitating more research on system security and model robustness.

One potential risk of our work is that the methodology may be used to launch an adversarial attack against online chat services or computer networks.
We believe the contribution of revealing the vulnerability and robustness of conversational models is more important than such risks, as the research community could pay more attention to different attacks and improves the system security to defend them.
Therefore, it is important to first study and understands adversarial attacks.

\section*{Acknowledgements}
This work was supported by NSF CNS 2135625, CPS 2038727, CNS Career 1750263, and a Darpa Shell grant.

\bibliography{main}
\bibliographystyle{acl_natbib}

\appendix

\section{Additional Settings and Results}
\label{sec:additional_results}

\textbf{Details of Victim Models.}
For DialoGPT, we use \textit{dialogpt-small} that contains 12 attention layers with 768 hidden units and 117M parameters in total. 
For BART, we use\textit{bart-base} that has 6 encoder layers together with 6 decoder layers with 768 hidden units and 139M parameters.
For T5, we use \textit{t5-small} that contains 6 encoder layers as well as 6 decoder layers with 512 hidden units and 60M parameters in total.

\begin{figure}[h]
    \centering    \includegraphics[width=0.43\textwidth]{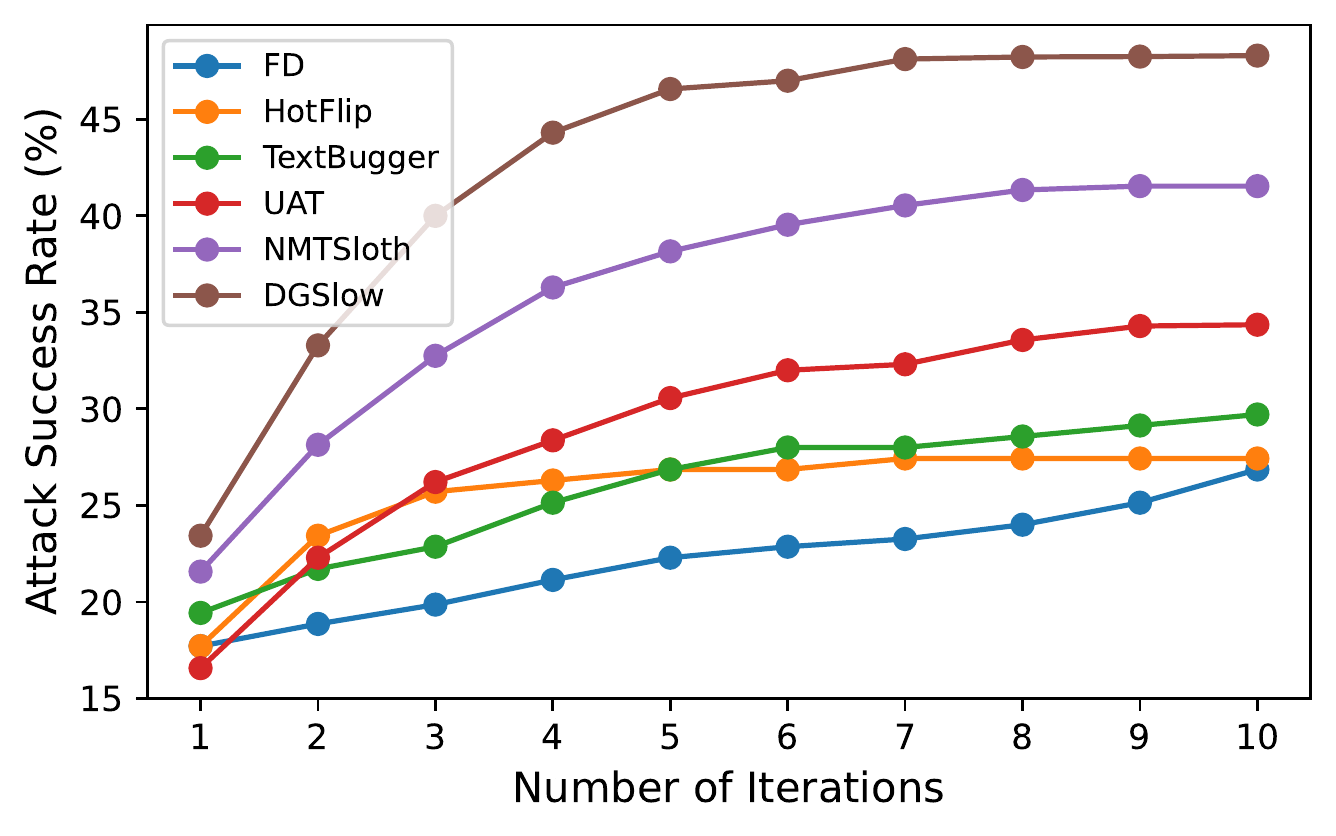}
    \caption{
    ASR vs. Number of iterations in BST when attacking BART.
    }
    \label{fig:asr_bart}
\end{figure}

\begin{figure}[h]
    \centering    \includegraphics[width=0.43\textwidth]{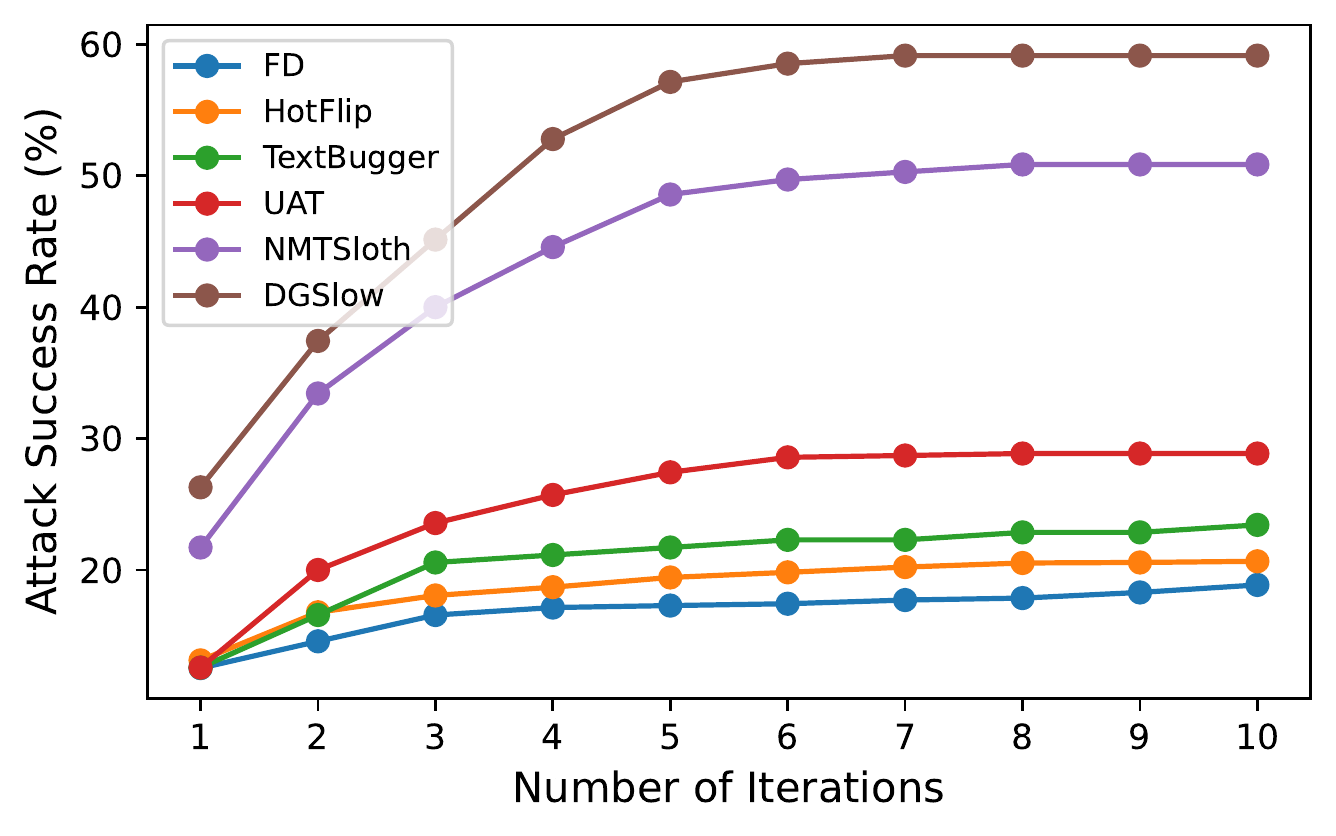}
    \caption{
    ASR vs. Number of iterations in BST when attacking T5.
    }
    \label{fig:asr_t5}
\end{figure}

\textbf{Attack Efficiency.} We evaluate the ASR under the restriction of iteration numbers for BART in Figure~\ref{fig:asr_bart} and T5 in Figure~\ref{fig:asr_t5}.
We observe that \Approach{} can significantly outperform all accuracy-based baseline methods.
Compared to the length-based NMTSloth, our method exhibits advantages when the iteration times goes large, showing the superiority of our adaptive search algorithm.

\textbf{METEOR Results.} We show the METEOR results for attacking the three models in four benchmark datasets in Table~\ref{tab:meteor}.
We observe that \Approach{} achieves overall the best METEOR scores, further demonstrating the effectiveness of our attack method.

\begin{table}[]
\centering
\resizebox{0.42\textwidth}{!}{
    \begin{tabular}{l|c|ccc}
    \hline
        \textbf{Dataset} & 
        \textbf{Method} &
        \multicolumn{1}{c}{\textbf{DialoGPT}} &
        \multicolumn{1}{c}{\textbf{BART}} &
        \multicolumn{1}{c}{\textbf{T5}} \\
        \hline
        \multirow{6}{*}{BST} & FD & 24.10 & 19.41 & 21.03 \\
        & HotFlip & 22.74 & 19.73 & 20.42 \\
        & TextBugger & 23.51 & 19.70 & 20.91 \\
        & UAT & 23.62 & 20.33 & 21.74 \\
        & NMTSloth & 23.15 & 22.03 & 19.52 \\
        & \Approach{} & \textbf{22.61} & \textbf{19.40} & \textbf{19.21} \\
        \hline
        \multirow{6}{*}{PC} & FD & 29.21 & 30.32 & 28.03 \\
        & HotFlip & \textbf{27.92} & 30.34 & 28.37 \\
        & TextBugger & 32.09 & 31.62 & 28.51 \\
        & UAT & 32.16 & 31.00 & 29.60 \\
        & NMTSloth & 29.04 & 31.51 & 27.39 \\
        & \Approach{} & 28.50 & \textbf{29.76} & \textbf{25.60} \\
        \hline
        \multirow{6}{*}{CV2} & FD & 8.13 & 11.14 & 9.53 \\
        & HotFlip & 9.42 & 11.71 & 9.50 \\
        & TextBugger & 8.91 & 10.82 & 9.13 \\
        & UAT & 9.84 & 11.53 & 8.67 \\
        & NMTSloth & 8.04 & 11.62 & 8.03 \\
        & \Approach{} & \textbf{8.00} & \textbf{10.52} & \textbf{7.71} \\
        \hline
        \multirow{6}{*}{ED} & FD & 11.06 & 11.03 & 11.04 \\
        & HotFlip & 9.82 & 13.42 & 10.53 \\
        & TextBugger & 11.92 & 10.43 & 10.23 \\
        & UAT & 11.87 & 11.93 & 10.11 \\
        & NMTSloth & 12.37 & 12.22 & 10.22 \\
        & \Approach{} & \textbf{9.66} & \textbf{9.70} & \textbf{9.91} \\
        \hline
    \end{tabular}}
    \caption{METEOR scores of attack methods on four datasets. \textbf{Bold} numbers mean the best metric values.}
    \label{tab:meteor}
\end{table}

\end{document}